# ADAPTING VEHICLE DETECTOR TO TARGET DOMAIN BY ADVERSARIAL PREDICTION ALIGNMENT

*Yohei Koga[1]\*, Hiroyuki Miyazaki[2], Ryosuke Shibasaki[2]*

[1]Independent researcher
[2]Center for Spatial Information Science, The University of Tokyo, Japan
\*Corresponding author, E-mail: monotaro333@gmail.com

## ABSTRACT

While recent advancement of domain adaptation techniques is significant, most of methods only align a feature extractor and do not adapt a classifier to target domain, which would be a cause of performance degradation. We propose novel domain adaptation technique for object detection that aligns prediction output space. In addition to feature alignment, we aligned predictions of locations and class confidences of our vehicle detector for satellite images by adversarial training. The proposed method significantly improved AP score by over 5%, which shows effectivity of our method for object detection tasks in satellite images.

## 1. INTRODUCTION

Recently, deep learning method have achieved state-of-the-art performance in various tasks, e.g., image classification, segmentation and object detection, and have been utilized in practical real-world applications. Vehicle detection in satellite images is one of such practical applications that often utilizes region-based object detectors [1]. In vehicle detection, region of interest is often different from training data in practice. This difference of data distribution between training data (source domain) and test data (target domain) causes severe performance degradation. To address this, domain adaptation (DA) method is applied. Typical DA methods find common feature space between source and target domains align the target domain features to the source domain and feed the aligned target domain features to a source domain classifier. While this methodology is substantially effective [2], there still remains subtle feature difference between source and target domains after feature alignment and the classifier cannot address the feature difference, which could hurt the performance.

In some way, a classifier needs to be adapted to target domain directly. Several previous works addressed this issue. [3] applied entropy minimization to their target domain classifier to urge the classifier to predict more confidently. [4] aligned prediction uncertainty of classifier outputs by adversarial training. Similarly, [5] aligned segmentation outputs among source and target domains in adversarial manner. In contrast to those normal DA approaches that align target domain classifiers to source domain, [6] aligned a source domain classifier to target domain by training a classifier using "synthesized" target domain features and achieved high accuracy in segmentation task. Although above-mentioned methods work well, they are all classification and segmentation tasks that fall into prediction of class confidences, thus are not necessarily directly applicable to object detection task. One difficulty is that not only class confidence but location need to be predicted in object detection task. In our preliminary experiments, entropy minimization did not improved performance. Likewise, [6] was not effective in our case. We conjecture that this was because an object detector needs to learn structured and sophisticated features useful for object detection on genuine labeled training data thus aligning source domain features to target domain could not achieve this objective.

To address the difficulty in object detection task, we propose a novel method of adapting a detector to target domain directly. We aligned predictions of both location and class confidence simultaneously utilizing adversarial training.

## 2. METHODOLOGY

To adapt a detector to target domain directly, we align target domain predictions to source domain. This is based on an assumption that source domain predictions are well-confident and quality ones because a source domain detector is trained on labeled dataset and this nature should be transferred to a target domain detector. We achieve this objective utilizing adversarial training that is applicable to not only class confidence but location prediction unlike entropy minimization.

### 2.1. Vehicle Detector

Considering accuracy, computational cost and simplicity, we adopted plain SSD [7] as our vehicle detector. SSD defines default boxes of several aspect ratios at every pixel in feature maps and a patch at every pixel is fed into a detector that consists of a location regressor and a classifier.

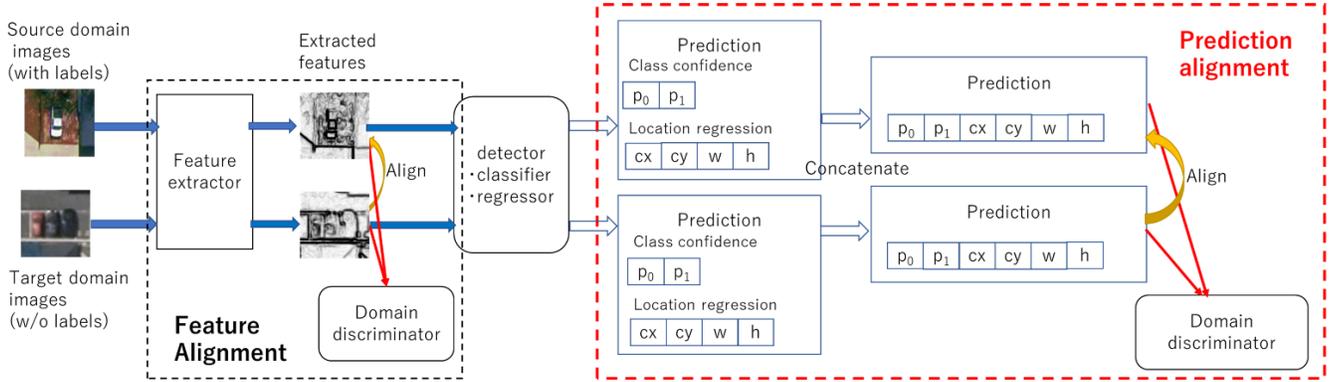

Fig. 1. The overview of the proposed prediction alignment.

Corresponding to one default box, the location regressor predicts location offsets from the default box ($cx$, $cy$) and ratios to the default box size ($w$, $h$), and the classifier predicts class confidence $P = (p_0, \cdots, p_c)$. $c$ represents a class number and $c = 0, 1$ in our case (0: background, 1: vehicle). The training objective is as follows:

$$\min_{M_s, F_s} L_{source} = \mathbb{E}_{(x_s, y_s) \sim (X_s \sim Y_s)} [L_{SSD}(F_s(M_s(x_s)), y_s)]$$

where $X_s$ and $Y_s$ are image and label examples drawn from the source domain distributions respectively; $M$ denotes a feature extractor and $F$ represents a detector that consists of the regressor and the classifier. We refer readers to the original SSD paper for detail of $L_{SSD}$.

**2.2. Prediction Alignment**

*2.2.1. Feature alignment*
As a basis of prediction alignment, we applied feature alignment by adversarial training. Without feature alignment, a model oscillated and did not converge. We regarded a local 3x3 patch at every pixel of a feature map as one feature unit and applied feature alignment to the unit because the unit is fed into the detector for one prediction. A discriminator is trained to predict domains of features and a feature extractor is trained to output undistinguishable features. Loss functions are as follows:

$$\min_{D_f} L_{feat\_dis} = -\mathbb{E}_{x_s \sim X_s}[\log D_f(M_s(x_s))]$$
$$- \mathbb{E}_{x_t \sim X_t}[\log(1 - D_f(M_t(x_t)))]$$
$$\min_{M_t} L_{feat\_ext} = -\mathbb{E}_{x_t \sim X_t}[\log D_f(M_t(x_t))]$$

where $X_t$ are image examples drawn from the target distribution; $D_f$ denotes the discriminator for feature alignment. In our proposed architecture, feature extractors of source and target domains are identical ($M_s(\cdot) = M_t(\cdot)$).

*2.2.2. Prediction alignment*
As shown in Fig. 1, we concatenated the outputs from the regressor and the classifier as a vector and applied adversarial training to the vector. A discriminator is trained to predict domains of predictions and the feature extractor and a detector are trained to output undistinguishable predictions. Loss functions are as follows:

$$\min_{D_p} L_{pred\_dis} = -\mathbb{E}_{x_s \sim X_s}[\log D_p(F_s(M_s(x_s)))]$$
$$- \mathbb{E}_{x_t \sim X_t}[\log(1 - D_p(F_t(M_t(x_t))))]$$
$$\min_{M_t, F_t} L_{pred\_det} = -\mathbb{E}_{x_t \sim X_t}[\log D_p(F_t(M_t(x_t)))]$$

where $D_p$ denotes the discriminator for prediction alignment. In this way, predictions of locations and class confidences are simultaneously aligned and consequently the quality of target domain prediction is improved.

*2.2.3. Class weight normalization*
In object detection task, area of foreground is usually only a small part of an entire image, thus most part is predicted as background. This extreme imbalance has negative effect on effective learning of prediction alignment. Although foreground objects need to be detected well-confidently in consequence of prediction alignment, gradients propagated from the predictions of background are dominant during training and predictions of foreground objects could not be aligned sufficiently. To address this, we introduced class weight normalization (CWN). We applied a weight inversely proportional to a number of examples of each class and balanced contribution of each class to prediction alignment as shown in Fig. 2. Specifically, we first calculated class weight as follows:

$$B = (b_0, \ldots, b_c), b_c = \frac{a_c N}{n_c C}$$

$n_c$ is a number of predictions where the highest class confidences were in class c, $N$ is a number of all predictions.

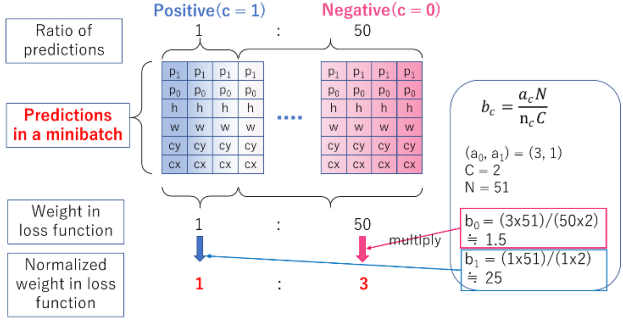

Fig. 2. The procedure of class weight normalization.

In our method, $N$ is a number of all predictions in one minibatch and $n_c$ is calculated every iteration. $a_c$ is a hyperparameter that defines the final weight of each class. If all $a_c$ are 1, each class is equally weighted.

Next, we allocated the class weight to each prediction according to highest class confidence as follows:

$$W = (w_0, ..., w_i), w_i = b_{argmax\, P_i}$$

$i$ is a number of default box in an image, $P_i$ is a prediction to $i$-th default box. Finally, $L_{pred\_dis}$ and $L_{pred\_det}$ can be updated to $WL_{pred\_dis}$ and $WL_{pred\_det}$ respectively.

*2.2.4. Training objective*
Overall training objective of our proposed prediction alignment is as follows:

$$\min_{D_f, D_p} L_{pred1} = L_{feat\_dis} + L_{pred\_dis}$$
$$\min_{M_s, M_t, F_s, F_t} L_{pred2} = L_{source} + L_{feat\_ext} + \alpha L_{pred\_det}$$

$\alpha$ is a coefficient that controls weight of $L_{pred\_det}$. In DA training, these two objectives are alternated.

## 3. EXPERIMENT

### 3.1. Dataset

We adopted COWC dataset [8] as source domain dataset. We only used RGB images. For target domain, we used aerial images of Tokyo, Japan. The images are processed by orthorectifications that are the same as the COWC dataset. Fig. 3 shows an image example. We processed these two raw image sets and obtained training and test datasets. Table 1 summarizes the specifications of datasets we used in this paper. The resampled resolution was 0.3m/pixel and we applied data augmentation (rotating images by 90, 180, and 270 degrees) to obtain training datasets.

### 3.2. Experimental Setting

*3.2.1. Implementation*
The input size 300x300 pixels. The backbone is VGG-16.

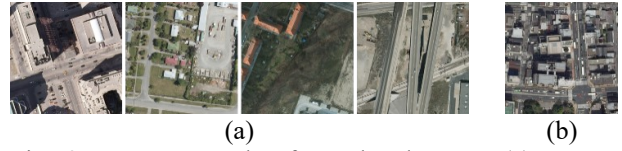

Fig. 3 Image examples from the datasets. (a) COWC dataset (source) and (b) aerial images of Japan (target).

Table 1. Specifications of the all datasets.

| Name | Domain | Usage | Size (pixel) | Image quantity | Vehicle annotation |
|---|---|---|---|---|---|
| Dataset S_train | Source | Training | 300 × 300 | 6,264 | 93,344 |
| Dataset T_train | Target | Training | 300 × 300 | 1,408 | - |
| Dataset T_test | Target | Test | 1000 × 1000 | 20 | 2,722 |
| Dataset T_labels | Target | Training | 300 × 300 | 1,564 | 23,024 |

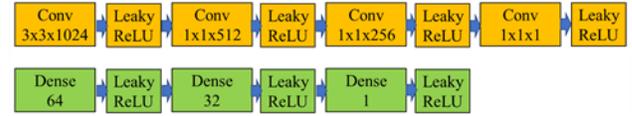

Fig. 4. Discriminator architectures. Orange: for feature alignment, green: for prediction alignment.

We changed default box sizes to {24, 30, 90, 150, 210, 270, 330} that is suitable to vehicle sizes in our datasets. We applied feature alignment and prediction alignment to only the shallowest feature map and predictions generated from that feature map that are mainly responsible for vehicle detection. For general object detection, all feature maps and predictions should be aligned. The discriminator architectures are shown in Fig. 4.

*3.2.2. Training configuration*
We first pretrained a vehicle detector on *Dataset S_train* and finetuned the model by DA methods on *Dataset T_train*. In the pretraining, we used the original configuration of SSD paper [7]. The batchsize was 32. Training iterations of pretraining were 40000 and the schedule of decaying learning rate was 28000 and 35000 iterations. In DA training, the batchsize was 64 that consisted of 32 source domain examples and 32 target domain examples. For comparison, we trained plain adversarial DA (only feature alignment) as a baseline for 15000 iterations and three variants of our proposed method: *w/o norm* for 15000 iterations, *norm D and P*, *norm P* for 10000 iterations. If not mentioned, we used $\alpha$ =1 and $(a_0, a_1) = (1,1)$. In *w/o norm*, we did not apply CWN. In *norm D and P*, we applied CWN to training of the discriminators, the feature extractor and the detector. Because we found applying CWN to discriminator training destabilized training and finally induced model collapse, we needed to use small $\alpha$ =0.1. In *norm P*, we applied CWN to only training of the feature extractor and the detector. Further, we adopt $(a_0, a_1) = (3,1)$ according to the hyper parameter k in the original SSD paper. Additionally, we trained a reference model using *Dataset S_train* and *T_labels*. We tested the obtained models on *Dataset T_test*. Experiments of DA methods were repeated 10 times and we report the statistics.

Table 2. Performance indicators of all methods.

| Method | Statistic | AP | F1 | PR | RR | FAR |
|---|---|---|---|---|---|---|
| Ref. | AVR | 78.7% | 80.9% | 83.7% | 78.4% | 15.3% |
|  | STDERR | - | - | - | - | - |
| w/o DA | AVR | 66.2% | 72.8% | 82.6% | 65.0% | 13.7% |
|  | STDERR | - | - | - | - | - |
| Plain adv | AVR | 74.7% | 80.5% | 88.3% | 74.0% | 9.9% |
|  | STDERR | 0.3% | 0.1% | 0.6% | 0.4% | 0.6% |
| w/o norm | AVR | 76.9% | 81.7% | 88.6% | 75.8% | 9.8% |
|  | STDERR | 0.3% | 0.1% | 0.4% | 0.4% | 0.4% |
| norm $D$ and $P$ | AVR | 79.0% | 80.6% | 81.0% | 80.8% | 20.1% |
|  | STDERR | 0.3% | 1.0% | 2.4% | 0.6% | 3.6% |
| norm $P$ | AVR | 79.8% | 82.2% | 84.5% | 80.2% | 14.8% |
|  | STDERR | 0.3% | 0.2% | 0.6% | 0.4% | 0.7% |

## 4. RESULT AND DISCUSSION

Results are shown in Table 2. The improvement of *w/o norm* from the baseline was slight. This indicates CWN was critical to performance improvement. Although *norm D and P* was substantially better than baseline, FAR was relatively high. While small $\alpha$=0.1 managed to train a model successfully, applying CWN to discriminator training seems to have been still too strong in our case. *norm P* was explicitly better than *norm D and P* and scored the best performance where AP was higher than the baseline over 5 points. Applying CWN to only training of feature extractor worked well in our experimental setting. However, this setting of CWN was determined heuristically and might not be the best for different scenarios. A versatile setting should be explored. Additionally, our method of CWN was simple and could be improved. These would be our future works.

While our proposed method is effective for object detection tasks of remote sensing images, there would be more difficult situations where resolution of labeled images is worse than test images. Even in such case, there are some works [9][10] that can be utilized in such situations and we can combine those solutions and our method. Further, our method could be extended to general object detection tasks. These would be explored in our future work.

## 5. CONCLUSION

We proposed a novel DA method called prediction alignment to address performance degradation caused by image feature difference that remains even after feature alignment. Our method significantly improved performance from the baseline and proved its effectiveness. While our CWN setting worked well, more general setting and a more sophisticated method of CWN should be explored. Further, our method should be extended to more difficult application scenarios including general object detection. These would be our future works.

## ACKNOWLEDGEMENT

The aerial images used in this paper were provided by NTT Geospace (http://www.ntt-geospace.co.jp/).